# A Modular Transradial Bypass Socket for Surface Myoelectric Prosthetic Control in Non-Amputees

Michael D. Paskett, Nathaniel R. Olsen, Jacob A. George, David T. Kluger, Mark R. Brinton, Tyler S. Davis, Christopher C. Duncan, and Gregory A. Clark

*Abstract*—Bypass sockets allow researchers to perform tests of prosthetic systems from the prosthetic user's perspective. We designed a modular upper-limb bypass socket with 3D-printed components that can be easily modified for use with a variety of terminal devices. Our bypass socket preserves access to forearm musculature and the hand, which are necessary for surface electromyography and to provide substituted sensory feedback. Our bypass socket allows a sufficient range of motion to complete tasks in the frontal working area, as measured on non-amputee participants. We examined the performance of non-amputee participants using the bypass socket on the original and modified Box and Block Tests. Participants moved 11.3 ± 2.7 and 11.7 ± 2.4 blocks in the original and modified Box and Block Tests (mean ± SD), respectively, within the range of reported scores using amputee participants. Range-of-motion for users wearing the bypass socket meets or exceeds most reported range-of-motion requirements for activities of daily living. The bypass socket was originally designed with a freely rotating wrist; we found that adding elastic resistance to user wrist rotation while wearing the bypass socket had no significant effect on motor decode performance. We have open-sourced the design files and an assembly manual for the bypass socket. We anticipate that the bypass socket will be a useful tool to evaluate and develop sensorized myoelectric prosthesis technology.

## I. INTRODUCTION

AS the field of upper-limb prostheses grows, functional tasks provide a crucial assessment in evaluating terminal devices and control algorithms [1]. With bypass sockets, researchers and developers can perform preliminary motor control testing and make improvements before formally testing with an amputee [2]. Although useful, derivative measures of control performance such as classifier accuracy, cross-movement error, or Fitts' law do not fully encapsulate a user's perception of control fidelity and performance [3]–[5]. Thus, early, in-the-loop testing uniquely informs iterative development of control algorithms and mimics upper-limb amputee use beyond what is possible with online/offline metrics.

Existing bypass sockets have limitations due to terminal device placement, restricted motion, and/or occlusion of the intact arm. Additionally, they are generally limited to a single terminal device [2], [6]–[14]. Exploring motor control through surface electromyography (sEMG) requires access to the skin [15], [16]. Researchers can also provide sensory feedback to the user through instrumented gloves, and this requires access to the intact hand [17].

We designed a bypass socket integrating three unique features absent in other bypass sockets: (1) it preserves access to the intact limb for sEMG recording from the forearm and sensory feedback on the hand, (2) it maintains the range of motion necessary for functional tasks, and (3) it easily adapts with various terminal devices and/or users. By combining these three features with minimal mass and mechanical complexity, we have created a "universal," open-source, modular bypass socket that can be readily adopted by other researchers.

## II. DEVICE DEVELOPMENT

### A. Criteria

#### 1) Preserve Access to the Intact Limb

Advanced control algorithms generally utilize a higher number of input channels for training compared with commercial myoelectric prostheses [16], [18]–[21]. With some exceptions (Coapt LLC, Chicago, IL), commercial myoelectric devices require two electrodes: one placed on the flexor mass and the other on the extensor mass of the forearm [22]. Using a multichannel array for sEMG requires more surface area, but provides more signals that can be used to provide intuitive

Manuscript submitted: 10/23/2018. This work was supported by NSF Award No. ECCS-1533649, and the Hand Proprioception and Touch Interfaces (HAPTIX) program administered by the Biological Technologies Office (BTO) of the Defense Advanced Research Projects Agency (DARPA), through the Space and Naval Warfare Systems Center, Contract No. N66001-15-C-4017. The content herein represents the authors' views, not necessarily their employers' or sponsors'.

M. D. Paskett, N. R. Olsen, J. A. George, D. T. Kluger, M. R. Brinton, and G. A. Clark are with the Department of Biomedical Engineering, University of Utah, Salt Lake City, Utah, USA, 84112 (emails: michael.paskett@utah.edu; nathanolsen2@gmail.com; jake.george@utah.edu; david.kluger@utah.edu; mark.brinton@utah.edu; greg.clark@utah.edu).

T. S. Davis is with the Department of Neurosurgery, University of Utah, Salt Lake City, Utah, USA, 84112 (email: tyler.davis@hsc.utah.edu).

C. C. Duncan is with the Department of Physical Medicine and Rehabilitation, University of Utah, Salt Lake City, Utah, USA, 84112 (email: christopher.duncan@hsc.utah.edu).

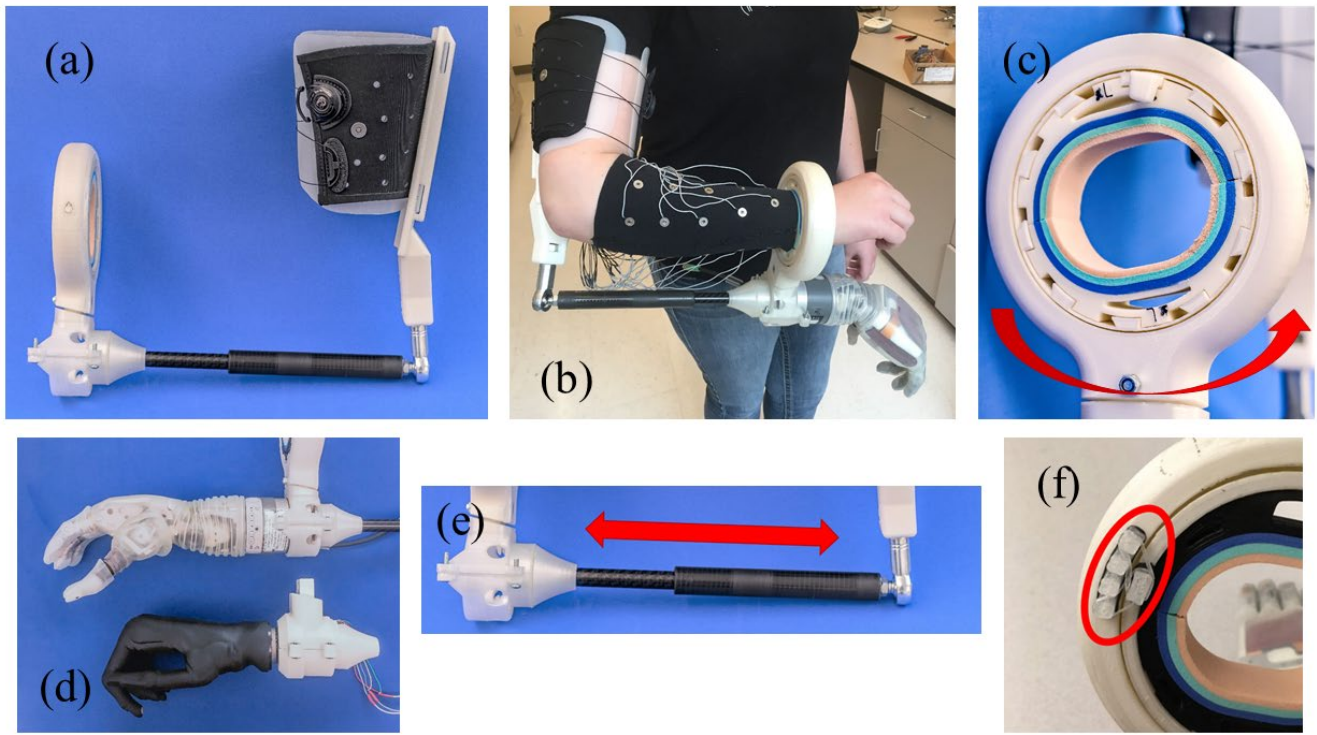

**Figure 1**. The bypass socket and Boa® attachment system (a) leaves forearm musculature free for surface electromyography by attaching at the upper arm and wrist (b). The wrist attachment allows unimpeded wrist flexion, extension, deviation, rotation, and is easily customized for any wrist size (c). Terminal devices are easily interchanged with the modular terminal device attachment (d; top: DEKA LUKE arm, bottom: i-limb™ ultra). The carbon fiber telescoping system accommodates changes in length requirements due to elbow flexion as well as various users (e). Optional ability to add elastic resistance to wrist rotation (f).

control of a high-degree-of-freedom prosthesis. Another consideration is the ability to perform sensorimotor integration experimentation, where the intact hand must be accessible for providing various forms of noninvasive vibrotactile or electrotactile substituted sensory feedback [17], [23], [24]. For these reasons, a bypass socket should preserve as much access to the intact limb as possible.

*2) Provide Free Motion in a Frontal Working Area*

Bypass sockets should enable the motion necessary for completing functional tasks [2], [8], [9], [12]. Although an ideal bypass socket allows full natural range of motion of the wrist and elbow, a reduction in mechanical complexity and distal mass favors a compromise in range of motion, while allowing the motion necessary for activities of daily living [25]. Most functional tasks of object manipulation and environmental interaction require a user to have freedom of motion in a frontal working area that heavily favors midrange joint angles of the elbow and wrist.

Bypass sockets tend to increase baseline activity in the sEMG signal, arising from added weight and restricted motion [26]. This increased baseline activity can be mitigated by including adequate range of motion and secure suspension while also reducing joint friction and overall distal mass. Correspondingly, transradial sockets produce a similar unavoidable change in baseline activity; therefore, complete elimination is both unnecessary and impractical. A bypass socket design should minimize added sEMG baseline activity while maintaining the motion necessary for completing functional tasks.

*3) Adaptability*

Existing bypass sockets generally support a single terminal device and are not easily adapted for other terminal devices [2], [7], [10], [13], [14]. This lack of adaptability is due to variable build lengths, distributions of distal mass, and heterogeneity of electromechanical attachments of the terminal device to the socket, reducing the number and type of compatible terminal devices. Conversely, modularity, facilitated by 3D printing, allows for rapid modifications to length and electromechanical attachments, reducing potential confounders. Lastly, because prostheses vary greatly in weight, a bypass socket should be able to support prostheses up to 2.3 kg, adequate for commercially available transradial prostheses [27]–[31].

*B. Design*

We designed our bypass socket to minimize mass, ensure robustness, and accommodate a broad range of users and terminal devices for functional task performance (Fig. 1). This expectation drove our design in terms of access to the intact limb, sturdiness, and ease-of-use. We were able to create a device that provides sufficient functionality to the user while accommodating variations in terminal devices, methods of control, and feedback modalities.

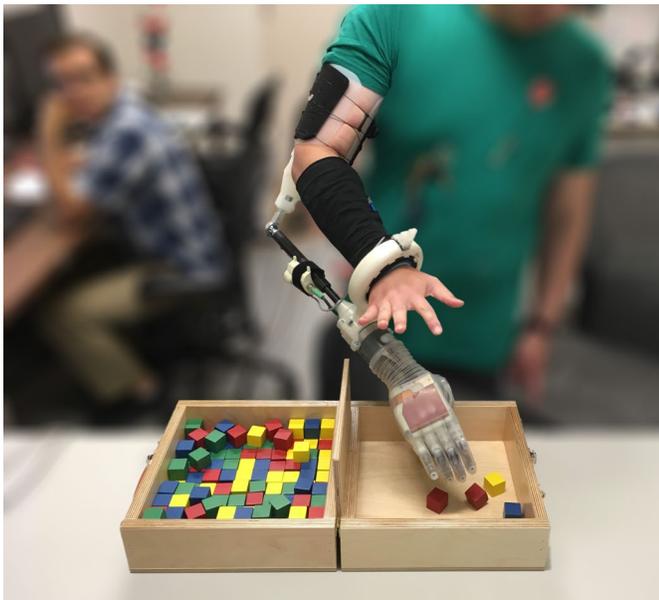

**Figure 2.** The Box and Block Test provides a quantified evaluation of a prosthetic system. With the bypass socket and DEKA LUKE arm on, participants transferred as many blocks as possible from one partition to another in 1 min. Participants also completed the modified Box and Block Test (not shown), which uses 16 blocks placed in a grid.

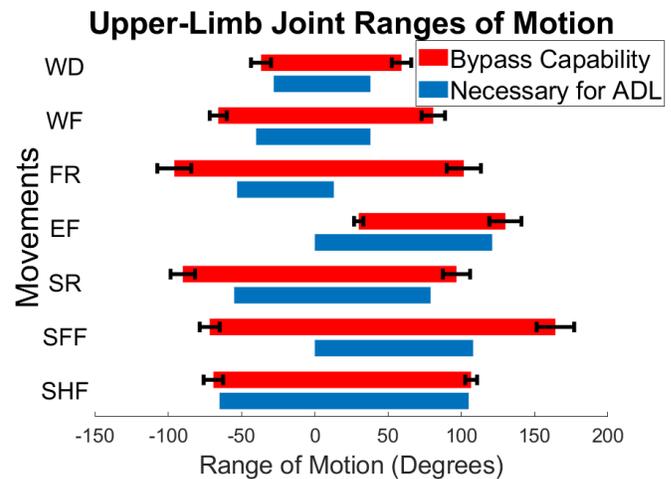

**Figure 3.** The bypass socket meets or exceeds the ranges of motion used in activities of daily living (ADLs) of most movement types as measured in Ref. [23] (mean ± SD).

WD: Wrist deviation. WF: Wrist flexion. FR: Forearm rotation. EF: Elbow flexion. SR: Shoulder rotation. SFF: Shoulder frontal flexion. SHF: Shoulder horizontal flexion.

*1) Preserve Access to the Intact Limb*

We maintained access to the intact limb by attaching in two places: at the upper arm and the wrist, distal to forearm musculature (Fig. 1a, 1b). The upper-arm attachment utilizes a Boa® system which provides ample support for heavy prostheses and supports a wide range of user sizes. The wrist attachment uses two 3D-printed pieces, which are joined around the wrist. A user's wrist can be locked in by sliding the wrist attachment pieces into the wrist component and rotating the attachment pieces relative to the internal locking mechanism. After this slight rotation, a "key" can be inserted between the attachment pieces and internal locking mechanism to prevent inadvertent removal (the process is visualized further in the open-source assembly manual linked in section VII). The two-point-of-contact arrangement enables the use of high-density sEMG to record from the extrinsic muscles driving finger and wrist movements. Our design also leaves the entire hand accessible, allowing experimenters to use various forms of mechanical or electrical haptic feedback on the exposed hand. The access to the hand allows for closed-loop sensory feedback experiments with a sensorized prosthetic. The contact area reduction of the bypass socket is an innovation that distinguishes our bypass socket from previous works [2], [7], [9].

*2) Provide Free Motion in a Frontal Working Area*

In the wrist, we preserved three degrees of freedom to minimize sEMG artifact due to flexion/extension, deviation, or pronation/supination hindrance (Fig. 1c). The wrist cuff is placed on the proximal carpal row allowing unimpeded flexion and extension of the wrist. The freely rotating design keeps the terminal device in the same relative location, volar to the user's wrist. This design choice optimizes object manipulation tasks and allows a two-degree of freedom prosthetic wrist to move in all planes unimpeded. Volar offset sacrifices the ability to carry or manipulate large objects (> 10 cm) with the palm facing up. We chose this offset with attention to the angle of the terminal device relative to the forearm, the distance between the user's hand and the terminal device, and the size of objects that can be handled with the palm facing upward.

A commercial ball joint was used at the elbow to preserve multiaxial freedom for the user. A telescoping system connects the ball joint and wrist attachment to accommodate both an increased length requirement during elbow flexion and users of various sizes (Fig. 1e). This combination of components provides the user sufficient range of motion for activities of daily living as compared with measurements from Ref. [25].

*3) Adaptability*

The bypass socket accommodates virtually any terminal device through modular terminal device attachments (Fig. 1d). The 3D-printed components can be easily modified in a parametric CAD software program, such as SolidWorks (SolidWorks Corporation, Waltham, MA), to fit a variety of terminal devices. At the time of publishing, the bypass socket supports the DEKA LUKE arm (DEKA, Manchester, NH) and Ottobock Quick Disconnect Wrist (Otto Bock HealthCare LP, Austin, TX).

The bypass socket design accommodates users of various sizes through interchangeable wrist attachments, the built-in telescoping system, and a Boa® cuff for the humeral attachment. The Boa® system (Click Medical, Colorado Springs, CO) provides both a sturdy and comfortable attachment for heavier terminal devices, such as the DEKA LUKE arm (1.5 kg; DEKA, Manchester, NH).

## III. Methods

### A. Signal Acquisition and Control Paradigm

Signal acquisition and techniques for decoding movement intent with a modified Kalman filter have been described previously [32], [33]. Briefly, sEMG was collected with the 512-channel Grapevine System (Ripple Neuro LLC, Salt Lake City, UT). 32 single-ended sEMG channels were acquired at 1 kHz and filtered with a 6th-order high-pass Butterworth filter (15 Hz), 2nd-order low-pass Butterworth filter (375 Hz), and 60, 120, and 180 Hz notch filters. Differential EMG signals for all 496 possible combinations were calculated, and all features (single-ended and differential) were sampled at 30 Hz using mean absolute value. The resulting 528 features were smoothed with an overlapping 300-ms boxcar filter.

Recordings were collected while the participants mimicked preprogrammed movements with the bypass socket donned (movements detailed in subsections B and C). Baseline activity of each feature was subtracted before training a modified Kalman filter. 48 channels were selected to be inputs for the modified Kalman filter based on a stepwise Gram-Schmidt algorithm [34]. The 48-channel subset and preprogrammed hand kinematics were used to fit the parameters of the modified Kalman filter. The output of the Kalman filter allowed participants to control a virtual or physical prosthetic hand with simultaneous and proportional control. The participants took part in the experiments after giving their informed consent, as approved by the University of Utah IRB.

### B. Functional Assessment of the Bypass Socket

Functional assessments, such as the Box and Block Test, provide quantified evaluations of a prosthetic system (Fig. 2). The Box and Block Test is a simple motor task in which a user transfers as many blocks as possible from one compartment of a box to the other, within 60 s. The Box and Block Test and modified Box and Block Test were completed five times each with naïve, non-amputee participants (n = 6 participants) wearing the bypass socket and an sEMG electrode sleeve (Fig. 1b). For the modified Box and Block Test, which uses 16 blocks placed in a grid, if a participant successfully moved all blocks within 60 s, their score was extrapolated to a rate for a full 60 s based on the completion time (e.g., if a participant moved all 16 blocks in 30 s, their score would be 32). Participants completed the tasks using a DEKA LUKE arm after training a modified Kalman filter for simultaneous and proportional control. Participants trained on three degrees of freedom: flexion/extension of thumb, index, and middle/ring/little fingers (which are mechanically coupled). The data for each degree of freedom consisted of five trials with 6.4 s duration (0.7 s moving toward and away from resting position, plus 5 s holding at maximum distance from resting position).

### C. Addressing Decode Stability During Wrist Rotation

A specific challenge with sEMG is that the electrodes shift relative to underlying musculature during user motion, which decreases decode performance. Our informal observation suggests this to be especially problematic during wrist rotation. To attempt to mitigate this effect, we added elastic bands to our rotating wrist attachment. We hypothesized that this elastic resistance to wrist rotation would provide two benefits: increase muscle activation with lesser rotation (minimizing electrode shift) and return the user's wrist to a neutral position when wrist movement is not intended. We chose not to fix the wrist attachment entirely, as doing so would cause user wrist rotations to change the volar offset of the terminal device.

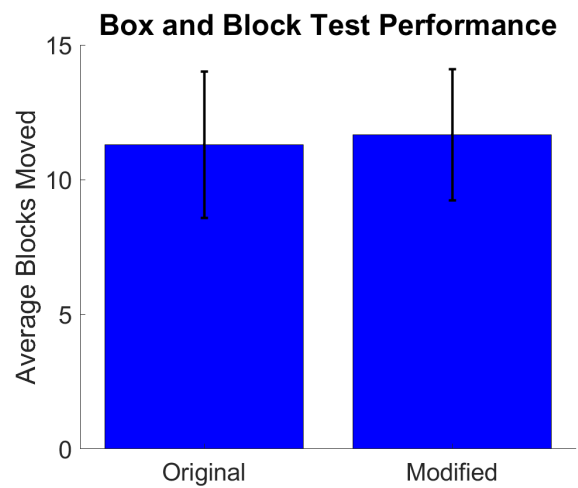

**Figure 4.** Box and Block Test performance of non-amputee participants with the DEKA LUKE Arm while wearing the bypass socket. Participants completed both the original and modified tests (n = 6 participants; mean ± SD).

We tested these hypotheses by comparing the performance of non-amputees with the wrist in a freely rotating and a resistance-added state (n = 11 participants). Initial testing was conducting on five participants, three of whom are co-authors and all had some experience with myoelectric prosthesis use. Further testing included six additional naïve participants. Resistance was added by attaching four 0.8 cm diameter elastic bands to the inner and outer portions of the rotating wrist attachment (Fig. 1f). With an elastic modulus of 0.4 N/mm$^2$ in the quasi-linear region (30-90 mm; roughly the elongation window of the bands during maximal wrist rotation), the bands provided approximately 0.34 N·m and 0.85 N·m of resistance at rest and maximal rotation, respectively.

Motor intent was decoded with a modified Kalman filter for simultaneous and proportional control using an sEMG sleeve. Participants trained on the six degrees of freedom available with the DEKA LUKE arm: flexion/extension of thumb, index, middle/ring/little (which are mechanically coupled), thumb adduction/abduction, wrist pronation/supination, wrist flexion/extension. The data for each degree of freedom consisted of four trials with 4.4 s duration (0.7 s moving toward and away from resting position, 3 s holding at maximum distance from resting position). Each participant completed one training set with elastic bands resisting wrist rotation, and one training set without elastic bands. This banded/free training configuration was alternated between participants to mitigate potential order effects. After completing both training sets, participants completed four virtual target matching tasks in an ABBA paradigm (e.g., one banded condition trial followed by two free condition trials, and finished with a second banded condition trial), where the user attempted to move the selected degree(s) of freedom to a target location. Targets were set at 50% of maximum distance. A single target set consisted of three movements (thumb flex, index flex, and middle flex) in three wrist orientations (neutral, pronation, and supination), for a total of nine movements (three digit-only and six combinations [digit + wrist]). One target trial consisted of a 0.7 s transition to the selected movement, a 5 s hold in the target position, and a 0.7 s transition back to resting position. An overall root-mean-square error (RMSE) was calculated from the target position and decoded position, similar to Ref. [32].

*D. Range of Motion Measurement*

Range of motion measurements of several upper-limb movements were taken to ensure adequate motion for a variety of activities of daily living. Participants donned the bypass socket, and a 12" goniometer was used to measure wrist deviation, wrist flexion, forearm rotation, elbow flexion, shoulder rotation, shoulder frontal flexion, and shoulder horizontal flexion (n = 6 participants). Participants were instructed to move the specified joint to the limit of the bypass socket or their physiological limits, whichever came first.

## IV. RESULTS

*A. Design*

Our bypass socket meets the established design criteria (Fig. 1). Specific design criteria are addressed in subsections 1-3.

*1) Preserve Access to the Intact Limb*

The Boa® attachment system ensures adequate suspension, minimizing distraction or rotation, and accommodates users with upper-arm circumferences between 20-35 cm. The modular wrist attachment pieces support user wrist circumferences up to 20 cm (Fig. 1c). These minimal attachment points leave the forearm clear for the use of sEMG and the hand clear for substituted sensory feedback.

*2) Provide Free Motion in a Frontal Working Area*

The bypass socket restricts motion very little, allowing sufficient ranges of motion for completing activities of daily living. We compared ranges of motion of six individuals wearing the bypass socket with the typical ranges of motion used when performing upper-limb activities of daily living, as measured previously (Fig. 3) [25]. Our results indicate motion exceeds activity-of-daily-living requirements for wrist deviation, wrist flexion, forearm rotation, shoulder rotation, shoulder frontal flexion, and shoulder horizontal flexion. Elbow extension is hindered by approximately 30 degrees, as the commercial ball joint limits full extension of the elbow. The design of the bypass socket keeps the terminal device in the same position relative to the user's wrist, regardless of wrist orientation, providing the user consistency in the location of the terminal device.

*3) Adaptability*

The overall design of the bypass socket provides versatility for both users and terminal devices. The modular, 3D-printed components enable researchers and developers to modify the bypass socket for a specific terminal device and switch between devices with ease. The Boa® upper-arm attachment securely supports heavier terminal devices, such as the DEKA LUKE arm. The Boa® system and 3D-printed wrist attachments adapt to virtually any user.

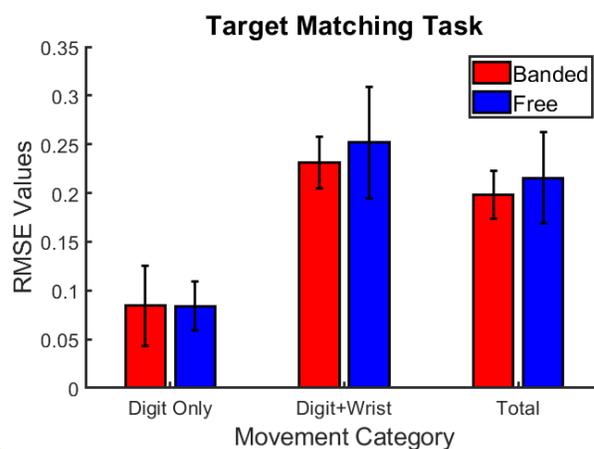

**Figure 5.** Results from a target matching task comparing wrist rotation with and without resistance from elastic bands (banded, free, respectively) indicate adding resistance to wrist rotation does not significantly alter decode performance (n = 11 participants). The "Total" movement category represents the overall RMSE when both categories (digit only and digit + wrist) are combined.

*B. Functional Assessment of the Bypass Socket*

An idealized bypass socket should mimic the characteristics and performance of a prosthetic socket and allow non-amputees to complete functional tasks similarly to amputees. Non-amputees were able to move 11.3 ± 2.7 blocks in the original Box and Block Test (mean ± SD; Fig. 4; n = 6 participants) using the bypass socket. In the modified Box and Block Test, non-amputees moved 11.7 ± 2.4 blocks (mean ± SD; Fig. 4; n = 6 participants). The results from the Box and Block Test and modified Box and Block Test lie within the range of previously published values using amputee subjects [35]–[43]. These results indicate the bypass socket allows users to complete functional tasks typically used for measuring prosthesis functionality.

*C. Addressing Decode Stability During Wrist Rotation*

In an attempt to reduce electrode shift during wrist rotation and improve the execution of other movements, we examined the effects of adding external resistance to wrist rotation (banded configuration), relative to performance in a free-wrist configuration. The effects of the banded and free-wrist configuration varied greatly among participants (n = 11 participants), although there were no significant differences overall. In initial experiments with five experienced users, the performance was better for digit-only movements in the banded configuration, compared with the free configuration (paired t-test, Holm-Bonferroni-corrected for multiple comparisons, $p < 0.05$, correction factor of 3). However, these differences were not maintained overall when six additional, naïve users were included (digit-only RMSE 0.111 ± 0.097, banded configuration; and 0.095 ± 0.044, free configuration; mean ±

SD; Fig. 5; $p > 0.05$). Similarly, there were no statistically reliable improvements overall in combination (digit + wrist) or total-movement RMSE values in the banded configuration (0.243 ± 0.046 and 0.214 ± 0.056 respectively), compared with the free configuration (0.252 ± 0.064 and 0.217 ± 0.056; means ± SDs; Fig. 5). Performance from one participant on digit-only movements and one participant on combination (digit + wrist) movements were removed as outliers, based on the 1.5 x interquartile range method for outlier detection [44].

## V. DISCUSSION

We developed a novel bypass socket that allows non-amputee participants to test and evaluate prosthetic systems and incorporated unique features not present in other bypass socket designs. Importantly, our bypass socket provides access to the intact limb for sEMG and is easily customizable for alternative terminal devices. The Box and Block Test and modified Box and Block Test demonstrated that the bypass socket can be used for functional tasks and provides an integrative solution for improving myoelectric prosthetic systems. Typically, researchers are limited to evaluating a single component of such a system (e.g., control algorithm, terminal device, feedback modality), which is incomparable to the evaluation of the system as a whole. Although the Box and Block Test does not fully extrapolate to prosthetic functionality, it is valuable as a broadly published metric in comparing the performance of non-amputee and amputee participants [2], [5], [45], [46]. We found the performance of naïve users with the bypass socket to be similar to the performance of amputee participants [35]–[43]. Although differences in terminal devices and control algorithms prevent a fair comparison with previous studies, the bypass socket seemingly provides a modest approximation of amputee usage.

Our implementation of a freely rotating wrist provides a unique feature: the terminal device maintains a volar offset relative to the user's wrist. This feature comes with a compromise due to the nature of sEMG recordings: wrist rotation causes surface electrodes to shift relative to the underlying musculature, which can negatively impact motor decode performance. To attempt to establish a balance between rotational freedom (which keeps the terminal device in a consistent location) and fixation (which reduces relative electrode shift), we attached elastics to the rotating aspect of the wrist attachment. Overall this manipulation did not provide benefits across the entire population; more experienced users showed positive benefits, but naïve users did not. Perhaps less experienced individuals have less stereotyped movements, which could negatively impact decode quality. Further work could include exploring potential performance differences between naïve and experienced individuals or finding an optimal resistance.

We were able to preserve natural range of motion, especially in a frontal working area as is required for functional assessments. The bypass socket does slightly restrict the full extension of the elbow (by ~30°); however, we have not found this to be a functionally important hindrance to task completion. Should a user need elbow extension beyond the current limit, a hinge could be added to the humeral attachment, proximal to the ball joint. The bypass socket is simple to set up for different users and can be donned or doffed in less than two minutes. It allows for future expansion to explore sensory substitution through an instrumented glove, thus creating a closed-loop system representative of the path of current technological thrusts [18], [47], [48].

Virtual reality provides another means of assessing a myoelectric prosthesis system. However, performance in the virtual realm does not always translate to performance in the physical world [49]. A virtual world must be parameterized in terms of physics, weight, friction, compliance, etc., whereas in the physical world, those properties are inherent. This mismatch between intrinsic natural properties and defined virtual properties can create a performance gap. Even without an additional feedback system, a bypass socket user generally feels collisions with physical objects through transmitted vibration, unlike the virtual realm. Thus, with current technologies, we see a bypass socket as more functionally relevant than a virtual environment for development and limited deployment using non-amputee participants.

## VI. CONCLUSION

Upper-limb prosthetic development has several distinct domains (e.g., control algorithms, sensory feedback modalities, terminal device development). Although these different elements are often evaluated in their respective areas correctly, prosthetic users judge the device holistically and indivisibly. Our bypass socket bridges these domains in a simple, versatile, and novel way, providing a holistic picture of the real-world performance of a prosthetic system. This work represents the first bypass socket to preserve access to the hand and forearm without encumbering user motion, and maintain adaptability for virtually any prosthetic device. These features make our bypass socket a valuable tool for researchers, clinicians, educators, and insurers.

## VII. OPEN-SOURCED

We have open-sourced our designs for the bypass socket. An instruction manual for creating the bypass socket is available at https://github.com/mpaskett/university-of-utah-bypass-socket. The GitHub repository contains all the necessary parts for 3D printing in both .stl and .sldprt file types. We hope the bypass socket can become an integral tool in evaluating upper-limb prostheses useable by multiple researchers.

## VIII. ACKNOWLEDGMENT

We thank Click Medical for help with the Boa® upper-arm attachment, as well as the University of Utah's Center for Medical Innovation for assistance with 3D printing.